\documentclass{pbmlarxiv}

\usepackage{epstopdf}
\usepackage{amsmath}
\usepackage{caption}
\usepackage{hhline}
\usepackage{booktabs}
\usepackage{hyperref}

\newcommand\blfootnote[1]{
  \begingroup
  	\renewcommand
    \thefootnote{}
    \footnote{#1}
    \addtocounter{footnote}{-1}
  \endgroup
}

\begin{document}

\title{Extracting Parallel Paragraphs from Common Crawl}

\institute{label1}{
  Charles University,
  Faculty of Mathematics and Physics,
  Department of Software Engineering
}
\institute{label2}{
  Charles University,
  Faculty of Mathematics and Physics,
  Institute of Formal and Applied Linguistics
}

\author{
  firstname=Jakub,
  surname=Kúdela,
  institute=label1
}
\author{
  firstname=Irena,
  surname=Holubová,
  institute=label1
}
\author{
  firstname=Ondřej,
  surname=Bojar,
  institute=label2,
  corresponding=yes,
  email={bojar@ufal.mff.cuni.cz},
  address={
    ÚFAL MFF UK (Linguistics)\\
    Malostranské náměstí 25\\
    11800 Praha\\
    Czech Republic
  }
}

\shorttitle{Extracting Parallel Paragraphs from Common Crawl}
\shortauthor{Kúdela, Holubová, Bojar}

\PBMLmaketitle

\begin{abstract}

Most of the current methods for mining parallel texts from the web assume that web pages of web sites share same
structure across languages. We believe that there still exists a non-negligible amount of parallel data spread across
sources not satisfying this assumption. We propose an approach based on a combination of bivec (a bilingual extension
of word2vec) and locality-sensitive hashing which allows us to efficiently identify pairs of parallel segments located
anywhere on pages of a given web domain, regardless their structure. We validate our method on realigning segments from
a large parallel corpus. Another experiment with real-world data provided by Common Crawl Foundation confirms that our
solution scales to hundreds of terabytes large set of web-crawled data.

% License: Creative Commons Attribution-NonCommercial-NoDerivatives 4.0 International Public License
\blfootnote{This work is licensed under a \href{https://creativecommons.org/licenses/by-nc-nd/3.0/}{CC BY-NC-ND 3.0} license.}

\end{abstract}

% The body of the article
% =======================
% The PBML class is modelled after the standard article class. This means
% that you can use almost everything that is allowed in articles as described
% in the textbooks of LaTeX. We support sectioning commands \section,
% \subsection and \subsubsection.

% In addition to the \cite command, you can use natbib style of citations.
% PLEASE, use \citet instead of \cite if the author names are part of the sentence.
%%\citet{PDT2} show \ldots \\ % This renders as "Hajič et al. (2006) show" instead of "(Hajič et al., 2006) show".
%%\ldots tree-based annotation (e.g.~\citealp{PDT2}).

% For figures and tables always use "figure" resp. "table" floating environments
% and always supply a caption (under the figure/table). See the paragraph about color images below.

\section{Introduction}
\label{sec:introduction}

The web is as an ever-growing source of considerable amounts of parallel data that can be mined and included in the
training process of machine translation systems. The task of \textit{bilingual document alignment} can be generally
stated as follows: Assume we have a set of documents written in two different languages, where a document is a plain
text of any length (a sentence, a sequence of multiple sentences, or even a single word). The goal of the task is to
collect all pairs of documents in different languages that are mutual translations of each other.

The majority of methods for bilingual document alignment assume that source documents are web pages and they rely on
their internal structure or structure of their URLs (e.g.~\citealp{Resnik:2003,Espla:2009}). By this filtering for
similar structure, we can lose a considerable amount of parallel data. Our proposed method is thus not based on page
structure comparison. Instead, we search for parallel paragraphs (or sentences) regardless their organization in the
page or in the web site. By moving to these finer units, we have to rely more on the actual content of the paragraphs.

To overcome the well-known problems of text data sparseness, we use bivec \cite{Luong:2015}---a bilingual extension of
currently popular word embedding model word2vec~\cite{Mikolov:2013}. To deal with the possibly large amount of input
documents, we make use of recently studied strategies for locality-sensitive hashing~\cite{Charikar:2002,Andoni:2008}.
Note that any finer alignment of the documents, such as sentence alignment (unless the documents consist of individual
sentences), is beyond the scope of our work. However, the methods for obtaining sentence alignment for a document-aligned
parallel corpus are well explored~\cite{Tiedemann:2011} and can be easily applied to the output of our method.

Related work has been described by~\citet{Roy:2016} and~\citet{Lohar:2016}. However, the discussed strategies use the
original monolingual word2vec model in contrast to our solution, which makes use of the bilingual bivec model to obtain
word vectors in a common vector space. Moreover, neither of the two approaches uses locality-sensitive hashing, which is
utilized in our method to achieve better speed performance with regard to scalability.

The rest of the paper is structured as follows: In Section~\ref{sec:proposed_method} we describe the proposed method.
Section~\ref{sec:experiments} provides the results of the experiments and Section~\ref{sec:conclusion_and_future_work}
concludes and outlines future work.

\section{Proposed Method}
\label{sec:proposed_method}

For our purposes, we refine the specification of bilingual document alignment. Let us assume we have a collection of
documents in two languages of interest, organized into \textit{bins}. Each bin holds two sets of documents, one in the
source language, the other in the target language, and represents a standalone set of input documents for the original
task. For each bin we want to find all the pairs of parallel documents (one in the source language, the other in the
target language) present within the bin.

A bin can contain up to millions of documents and it is not required to have a balanced language distribution.
Individual bins may vary in size. The binning is a way to restrict the set of considered pairs. No pairs are aligned
across different bins, nor between the documents of the same language. The smaller the size of a bin, the better the
quality of the resulting alignment (because fewer document pairs need to be considered). It also takes less time and
memory to align a smaller bin.

When mining bilingual parallel corpora from the web, we can form a bin for every identified bilingual web domain, simply
by taking all paragraphs in the two languages languages of interest, scraped from the web domain. We could be less
permissive and create bins spanning several web domains, but at this moment we are leaving this option for the future.

\subsection{Training Part I: Bilingual Dictionary, Bilingual Word Vectors}
\label{sec:training_part_1}

\begin{figure}[!htb]
  \centering
  \begin{minipage}[b]{0.28\textwidth}
    \includegraphics[width=\textwidth]{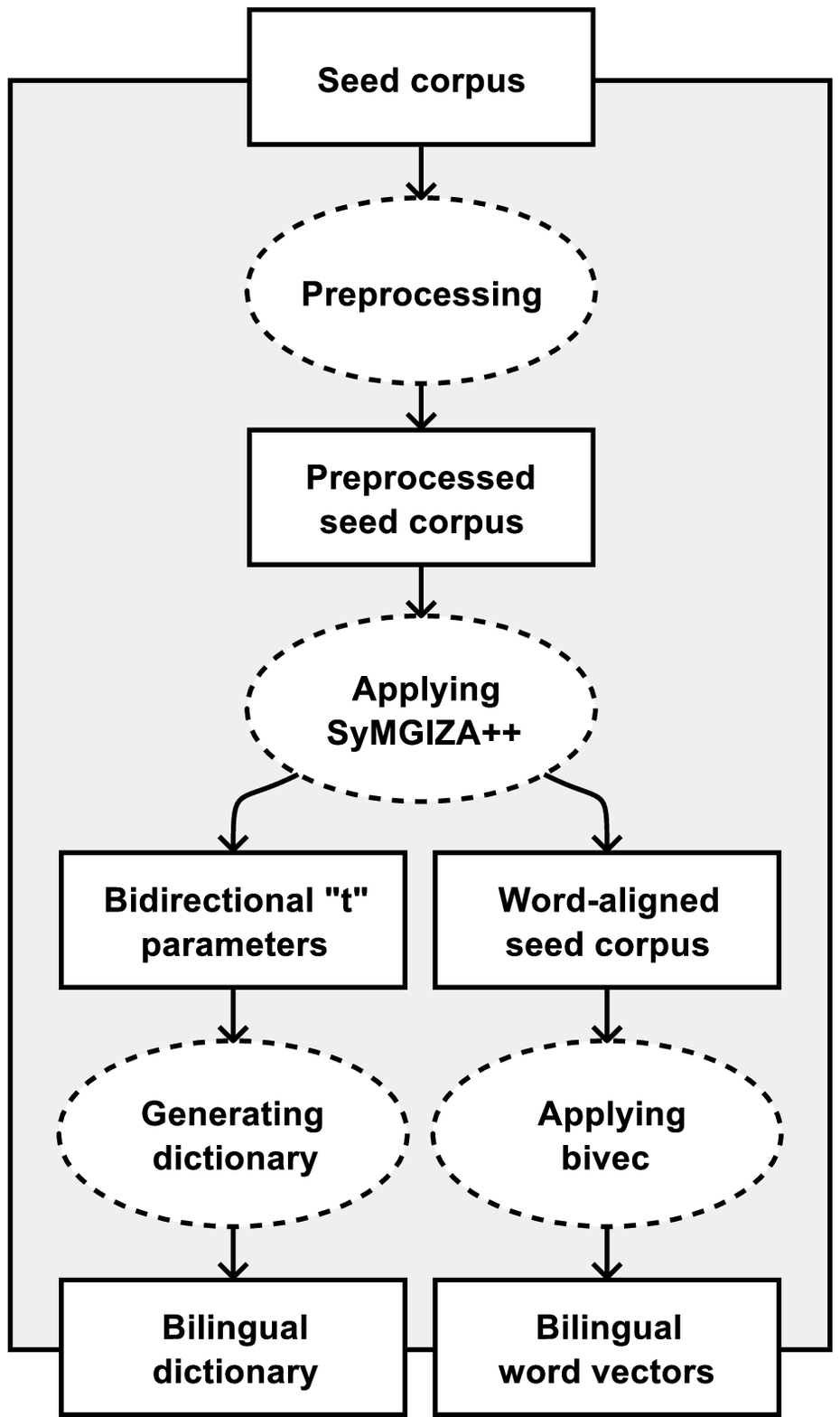}
    \captionsetup{justification=centering}
    \caption{Method:\\Training part I}
    \label{fig:method_1}
  \end{minipage}
  \hfill
  \begin{minipage}[b]{0.35\textwidth}
    \includegraphics[width=\textwidth]{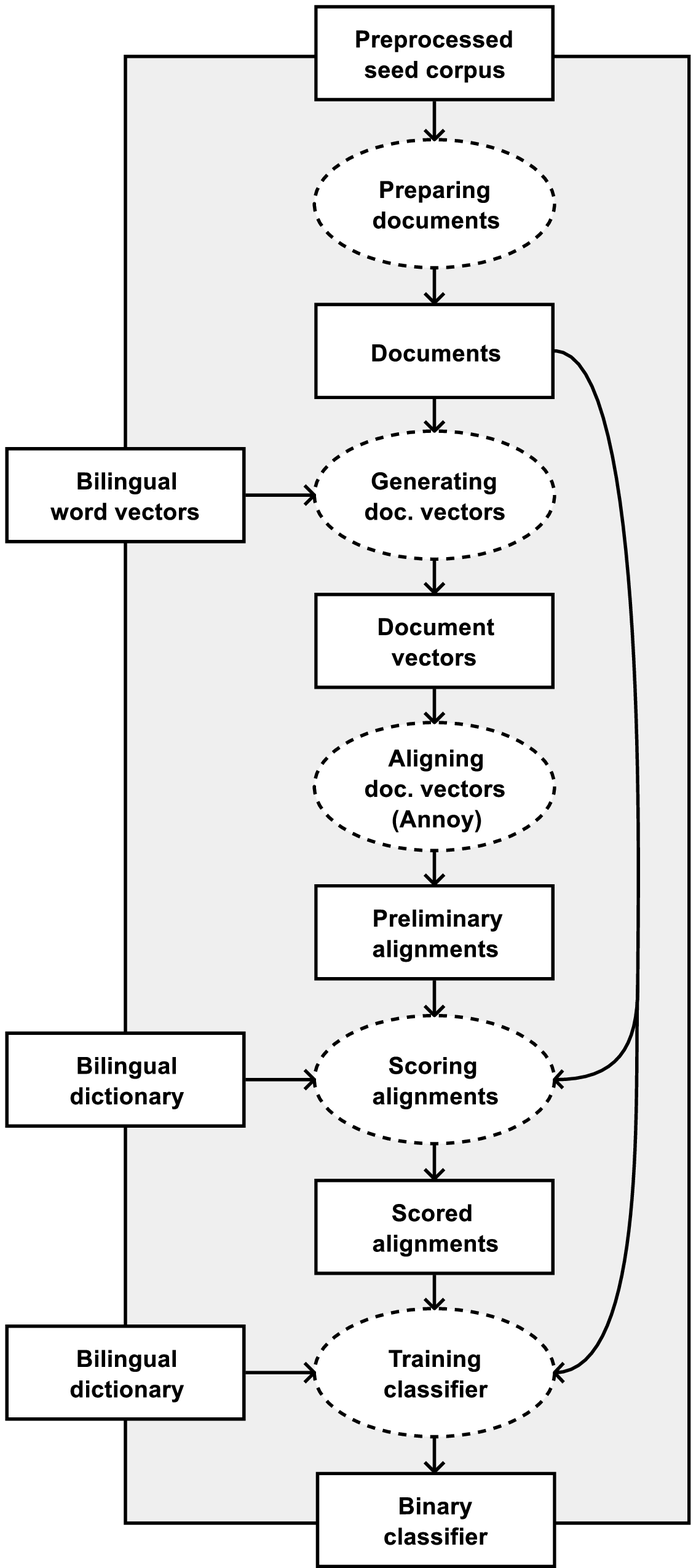}
    \captionsetup{justification=centering}
    \caption{Method:\\Training part II}
    \label{fig:method_2}
  \end{minipage}
  \hfill
  \begin{minipage}[b]{0.35\textwidth}
    \includegraphics[width=\textwidth]{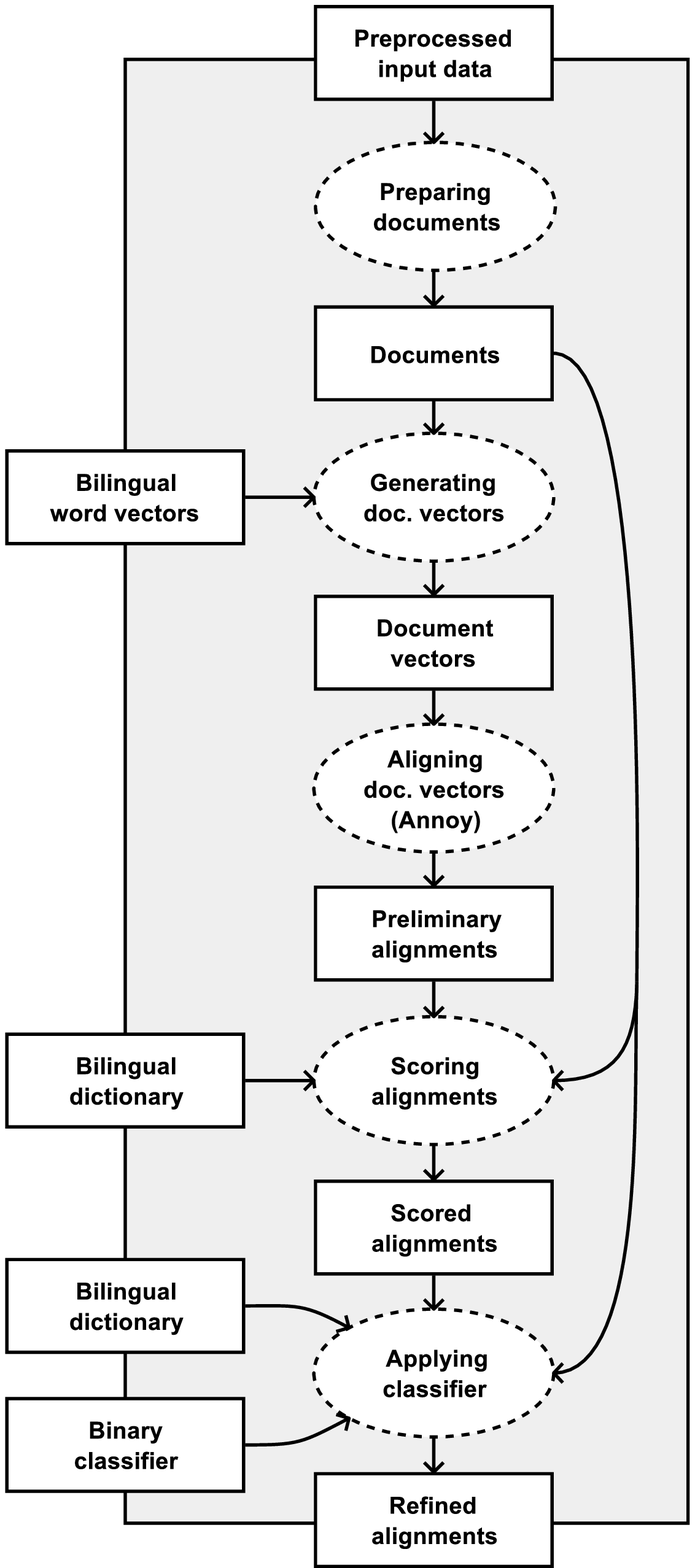}
    \captionsetup{justification=centering}
    \caption{Method:\\Application}
    \label{fig:method_3}
  \end{minipage}
\end{figure}

The proposed method is supervised and needs to be trained on an already existing sentence-aligned training parallel
corpus (i.e.~\textit{seed corpus}) for the language pair we are interested in. For better clarity, we distinguish
between two parts of the training process. This section describes the first part, which is depicted in
Figure~\ref{fig:method_1}. The objective of this part is to preprocess the seed corpus and create a bilingual dictionary
together with bilingual word vectors.

\subsubsection{Preprocessing Seed Corpus}
\label{sec:preprocessing_training_parallel_data}

The preprocessing may involve tokenization, lemmatization, stemming, truecasing, lowercasing, removing stop words, etc.
For individual language pairs, different preprocessing steps might help to gain better results. It is very important
that any type of preprocessing done to the seed corpus needs to be also applied to the input data before the alignment
process starts.

\subsubsection{Applying SyMGIZA++}
\label{sec:applying_symgiza}

The resulting corpus from the previous step is further cleaned by removing all such pairs where one of the sentences
contains more than $50$ tokens or does not contain any letter from any alphabet. Then SyMGIZA++~\cite{Junczys:2012} is
executed to obtain a word alignment for the preprocessed and cleaned seed corpus. This step includes preparation of word
classes and word co-occurrences which are used in the word alignment process.

SyMGIZA++ is a tool for computing symmetric word alignment models. It is an extension of MGIZA++~\cite{Gao:2008}, which
is in turn a successor of the historically original program called GIZA++~\cite{Och:2003}.

\subsubsection{Generating Dictionary}
\label{sec:generating_dictionary}

The bilingual dictionary is built using the final IBM Model ``t'' parameters estimated by SyMGIZA++. Each word pair
present in both directions having the harmonic mean of the ``t'' parameters (i.e.~\textit{weight}) over a certain
threshold is included into the dictionary together with the calculated weight.

\subsubsection{Applying bivec}
\label{sec:applying_bivec}

Word embedding is a common name for a set of techniques mapping words or phrases from a vocabulary to distributed word
representations in the form of vectors consisting of real numbers in a high-dimensional continuous space. Our method
takes advantage of bivec---a bilingual extension of word2vec. It creates bilingual word representations when provided
with a word-aligned parallel corpus.

The original word2vec is a group of models producing monolingual word embeddings. These models are implemented in form
of neural networks. They are usually trained to reconstruct the contexts of words. A monolingual corpus is needed for the
training. The training algorithm iterates over the words in the corpus while considering the surrounding words to be the
context of the current word. One word2vec model, called continuous bag-of-words (CBOW), is trained to predict the word when
given its context (without the word). Another model, named skip-gram, is trained to predict the context of a given word.

The authors of bivec proposed an extension of the original skip-gram model in the form of a joint bilingual
model---bilingual skip-gram. When trained, this model can predict the context of a given word in both languages. In
order to train, bivec requires a sentence-aligned parallel corpus and its word alignment. In fact, the word alignment is
not strictly necessary, and if it is not provided, the system uses a simple heuristic. However, with the alignment
provided, the results are better.

Our method follows the training by further processing the word-aligned seed corpus produced by SyMGIZA++ according to
the recommendations of bivec. All the sequences of numbers are replaced with a unified placeholder (the symbol ``\texttt{0}'')
and all the unknown symbols (e.g.~non-printable Unicode characters) with the specially dedicated \texttt{<unk>} tag.

With the word-aligned seed corpus processed, bivec is executed to create the bilingual word vectors (i.e. embeddings)
with $40$ dimensions. These vectors are known to have a greater cosine similarity for context-related words, even
cross-lingually. Although the number of dimensions is an unrestricted parameter, there is a reason why we keep it
relatively low. The word vectors are used to calculate the aggregate document vectors with the same number of
dimensions. The document vectors are then indexed using Annoy\footnote{\url{https://github.com/spotify/annoy}}---an
implementation of approximate nearest neighbors search. The documentation of Annoy suggests that it works best with the
number of dimensions less than $100$. On the other hand, the authors of bivec conducted the tests using $40$, $128$,
$256$ and $512$ dimensions. We have decided to use the only number of dimensions suitable for Annoy that has been tested
with bivec.

\subsection{Training Part II: Binary Classifier}
\label{sec:training_part_2}

The second part of the training process is illustrated in Figure~\ref{fig:method_2}. In this part, the method attempts to
find candidate sentence pairs by realigning the preprocessed seed corpus. This is performed by employing the bilingual
word vectors and locality-sensitive hashing. While working with the seed corpus, we know which of the candidate sentence
pairs are correct and we exploit this knowledge to train a binary classifier. The trained classifier is then used when
applying the trained method on the input data.

\subsubsection{Preparing Documents}
\label{sec:preparing_documents}

The method splits all the pairs of sentences from the seed corpus (i.e.~documents) into a set of equally large bins. The
last bin can be an exception. The size of a bin should be an estimate of its expected size in a real-world use case. In
our experiments, we split the corpus into training bins consisting of $50,000$ pairs of parallel documents, i.e.~$100,000$
individual documents. We believe that this amount of documents is a good upper-bound estimate of the total number of
paragraphs in either of the two languages located on a typical bilingual web domain.

\subsubsection{Generating Document Vectors}
\label{sec:generating_document_vectors}

For each document, an associated vector is generated using the bilingual word vectors obtained in the first part of the
training. To calculate the document vector, we utilize the tf-idf (i.e.~term frequency - inverse document frequency)
weighting scheme. For every unique document $d=(d_1, d_2, \ldots, d_n)$ a vector is generated as:

\begin{align}
  \operatorname{doc\_vector}(d)=\sum\limits_{i=1}^{n} \operatorname{tf-idf}(d_i, d) \times \operatorname{word\_vector}(d_i)
\end{align}

\noindent where $\operatorname{tf-idf}(d_i, d)$ is a tf-idf of the term $d_i$ in the document $d$ and
$\operatorname{word\_vector}(d_i)$ is the bivec word vector for the term $d_i$. If a word vector does
not exist for a given term, a zero vector is used instead.

In a smaller-scale experiment performed with our method, the tf-idf weighting scheme was compared to a plain sum
of the word vectors with an equal weight. The method yielded comparatively better results using the tf-idf scheme.

\subsubsection{Aligning Document Vectors (Annoy)}
\label{sec:aligning_document_vectors}

For each bin, independently of other bins, the following procedure is performed. A search index is built containing the
vectors of all the documents in the target language. To build the search index, the method uses Annoy operating with the
angular distance. Then, for every document in the source language, the index is searched to obtain $k$ approximate
nearest neighbors to its vector. This returns a list of candidate parallel documents in the target language to the
document in the source language. We call them \textit{preliminary alignments}.

Annoy is an implementation of approximate nearest neighbors search. Unlike the exact search methods, it does not
guarantee to find the optimum, but in many cases it actually does. This relaxation enables it to require less resources.
In particular, it needs less time, which becomes useful when dealing with larger amounts of data. Internally, Annoy uses
random projections to build up a forest of search trees---an index structure for the searching process. The algorithm
for building the index uses SimHash, which is a locality-sensitive hashing method.

\subsubsection{Scoring Alignments}
\label{sec:scoring_alignments}

Within the preliminary alignments, the top candidates are not necessarily the optimal ones. Therefore, our method
applies a scoring function to reorder the candidates creating the \textit{scored alignments}. This increases the
probability of the optimal documents to appear higher in their candidate lists. Given the document $d=(d_1, d_2,
\ldots,d_n)$ and its candidate document pair $c=(c_1, c_2, \ldots, c_m)$, where $d_i$ and $c_i$ are the individual
terms in the respective documents, the scoring function is defined as:

\begin{align}
  \operatorname{score}(d, c) = \operatorname{length\_similarity}(d, c) \times \operatorname{weight\_similarity}(d, c)
\end{align}

\noindent Both the functions $\operatorname{length\_similarity}(d, c)$ and $\operatorname{weight\_similarity}(d, c)$
have the range of $[0, 1]$. The idea is that the higher the result they return, the greater the possibility that the
pair is parallel. The $\operatorname{length\_similarity}(d, c)$ function compares the ratio of the documents' lengths.
It is based on the probability density function of the Gaussian (normal) distribution:

\begin{align}
  \operatorname{length\_similarity}(d, c)=e^{-\dfrac{\left(\frac{\operatorname{length}(c)}{\operatorname{length}(d)}-\mu\right)^2}{2\sigma^2}}
\end{align}

\noindent where $\frac{\operatorname{length}(c)}{\operatorname{length}(d)}$ is the actual ratio of the documents' lengths
(i.e.~total number of characters) and $\mu$ is the expected ratio with the standard deviation $\sigma$. The expected ratio
with its standard deviation can be estimated using the pairs of parallel sentences from the preprocessed seed corpus. The
other function $\operatorname{weight\_similarity}(d, c)$ is based on the IBM Model 1~\cite{Brown:1993} and uses the bilingual
dictionary created in the first part of the training. It is defined as:

\begin{align}
  \operatorname{weight\_similarity}(d, c)=\prod\limits_{i=1}^{n} \sum\limits_{j=1}^{m} \frac{\operatorname{weight}(d_i, c_j)}{m}
\end{align}

\noindent where $\operatorname{weight}(d_i, c_j)$ is the weight of the word pair $\langle d_i,c_j \rangle$ provided by
the dictionary if the word pair entry exists, otherwise it equals $10^{-9}$ (i.e.~``null weight'').

\subsubsection{Training Classifier}
\label{sec:training_classifier}

We use a binary classifier to decide whether to accept a proposed pair of documents as parallel or not. The chosen model
for the classifier is a feed-forward neural network trained by back-propagating errors~\cite{Rumelhart:1986}. The method
uses an implementation provided by PyBrain~\cite{Schaul:2010}. The classification is based on $4$ features. All of these
features have the range of $[0, 1]$. Given the document $d=(d_1, d_2, \ldots, d_n)$ and its candidate document pair
$c=(c_1, c_2, \ldots, c_m)$, the following text describes all the features.

The first feature $\operatorname{length\_similarity}(d, c)$ has been already defined in
Section~\ref{sec:scoring_alignments}. It scores the ratio of the documents' lengths against the expected ratio. The
second feature $\operatorname{length\_confidence}(d, c)$ provides a supplementary information for the first one, which
is neither reliable, nor effective when scoring pairs of short documents; however, it is substantial when comparing
pairs of long documents:

\begin{align}
  \operatorname{length\_confidence}(d, c)=1 - e^{-0.01 \times \operatorname{length}(d)}
\end{align}

\noindent This is a monotonically increasing function providing the model with an information of absolute length of
the document $d$. The higher the $\operatorname{length\_confidence}(d, c)$ is, the more authoritative the score of the
$\operatorname{length\_similarity}(d, c)$ should be deemed.

The third feature $\operatorname{weight\_similarity_2}(d_i, c_j)$ is a modified version of the one already defined
(i.e.~$\operatorname{weight\_similarity}(d, c)$). The original version was tested for the purposes of the
classification, but the results were poor. The ineffectiveness could be caused by the fact that the original function
lacks some proper normalization with respect to documents' sizes and returns extremely small values when comparing
larger documents. The modified version is defined as follows:

\begin{align}
  \operatorname{weight\_similarity_2}(d, c) = \frac{\sum\limits_{i=1}^{n} \operatorname{length}(d_i) \times
  \max\limits_{j=1}^{m}\left(\operatorname{weight_2}(d_i, c_j)\right)}{\sum\limits_{i=1}^{n} \operatorname{length}(d_i) \times
  \operatorname{sgn}(\max\limits_{j=1}^{m}\left(\operatorname{weight_2}(d_i, c_j)\right))}
\end{align}

\noindent where $\operatorname{length}(d_i)$ is the length of the term $d_i$ and $\operatorname{weight_2}(d_i, c_j)$
is defined as the weight of the word pair $\langle d_i,c_j \rangle$ provided by the dictionary if the entry exists;
however, if the entry does not exist and the two words are identical then it equals $1$, otherwise it returns $0$.

Let us explain the reason for the heuristic of $\operatorname{weight_2}(d_i, c_j)=1$ for a pair of identical words not
having an entry present in the dictionary. The same set of features is used when applying the trained method on the
input data. At that moment, occurrences of new words or special terms (e.g.~URLs or email addresses) are expected. The
heuristic considers a pair of identical words to be a perfect translation only if the dictionary does not contain other
relation.

Moreover, the weights are multiplied by the lengths of words due to an assumption that longer words are
usually less frequent, carry more meaning, therefore are more important for the sentence. The definition of
$\operatorname{weight\_similarity_2}(d, c)$ is an arithmetic mean of strongest relations between a source
word from $d$ and any of the target words from $c$, weighted by the lengths of source words. We can interpret
$\operatorname{weight\_similarity_2}$ as: ``Given our incomplete word translation dictionary, how likely are
these two documents parallel?''

The last feature $\operatorname{weight\_confidence_2}$ supplements the third one and it can be interpreted as: ``To what
extent does the dictionary cover the pairs of words in these documents?''. The formal definition is the following:

\begin{align}
  \operatorname{weight\_confidence_2}(d, c) = \frac{\sum\limits_{i=1}^{n} \operatorname{length}(d_i) \times
  \operatorname{sgn}(\max\limits_{j=1}^{m}\left(\operatorname{weight_2}(d_i, c_j)\right))}{\sum\limits_{i=1}^{n} \operatorname{length}(d_i)}
\end{align}

The process of training of the binary classifier starts by creating a supervised data\-set using the scored alignments.
For every document in the source language and its top candidate in the target language, a pair of input$\rightarrow$output
vectors is added into the supervised dataset as follows:

\begin{align}
  \begin{pmatrix}
    \operatorname{length\_similarity}(d, c) \\
    \operatorname{length\_confidence}(d, c) \\
    \operatorname{weight\_similarity_2}(d, c) \\
    \operatorname{weight\_confidence_2}(d, c)
  \end{pmatrix} \rightarrow \begin{cases}
    \begin{pmatrix} 0 \\ 1 \end{pmatrix} & \begin{matrix} \text{ if } \langle d,c \rangle \text{ are parallel} \end{matrix} \\\\
    \begin{pmatrix} 1 \\ 0 \end{pmatrix} & \begin{matrix} \text{ otherwise} \end{matrix} \\
  \end{cases}
\end{align}

\noindent The input vector consists of the 4 defined features, while the output vector encodes whether the documents
$\langle d,c \rangle$ are parallel or not. The first value of the output vector represents the probability of the documents
to be non-parallel. The second value is complementary to the first one. Before the network is trained, the collected supervised
dataset is subsampled to contain an approximately equal number of items representing parallel and non-parallel document pairs.
This helps the network to be less biased by the ratio of parallel and non-parallel pairs in the supervised dataset. At this
moment, it is also possible to reduce the size of the dataset to shorten the time it takes to complete the training.

For completeness, let us describe the configuration of the network. It has 4 input, 16 hidden and 2 output neurons. The
hidden neurons are arranged in a single layer. The input neurons are linear, the hidden layer uses the sigmoid function
and the output layer uses the softmax function.

\subsection{Application}
\label{sec:application}

The process of applying the trained method on the input data is illustrated in Figure~\ref{fig:method_3}. It is almost
identical with the procedure of the second part of the training (described in Section~\ref{sec:training_part_2}). Due to
this similarity, the following text does not cover the shared parts.

The input documents have to be preprocessed in the same way as the seed corpus during the training. Then, the
preprocessed documents have to be split into bins. When aligning paragraphs from the web, a bin can contain all the
paragraphs for both the languages scraped from one bilingual web domain. In this scenario, the names of the web domains
can be used as bin identifiers.

\subsubsection{Applying Classifier}
\label{sec:applying_classifier}

With the input dataset prepared, the process follows with the same steps as in the second part of the training. First,
vectors are generated for all the documents. Then, the document vectors are aligned by searching for nearest neighbors
of all the documents in the source language, resulting in preliminary alignments. In the final step, the trained
classifier is used to obtain \textit{refined alignments}. For every document in the source language and its top
candidate in the target language the trained network is activated in the same way it has been trained. The second value
of the output vector represents the confidence that the two documents are parallel. If the confidence is greater than a
user-defined threshold, the document pair ends up in the resulting refined alignments (i.e.~\textit{extracted corpus}).

\section{Experiments}
\label{sec:experiments}

Our experiments are solely focused on the Czech--English language pair. The first experiment is carried out using CzEng
1.0~\cite{Bojar:2012}---a Czech--English sentence-aligned parallel corpus. By realigning the CzEng 1.0 corpus, we can
evaluate the quality of the results automatically. The second experiment uses more realistic and noisy data provided by
Common Crawl Foundation.\footnote{\url{http://commoncrawl.org/}} The organization produces and maintains an open
repository of web-crawled data that is universally accessible and analyzable.

\subsection{Prealigned Data (CzEng 1.0) Experiment}
\label{sec:prealigned_data_experiment}

\begin{figure}[!htb]
  \centering
  \begin{minipage}[b]{0.28\textwidth}
    \includegraphics[width=\textwidth]{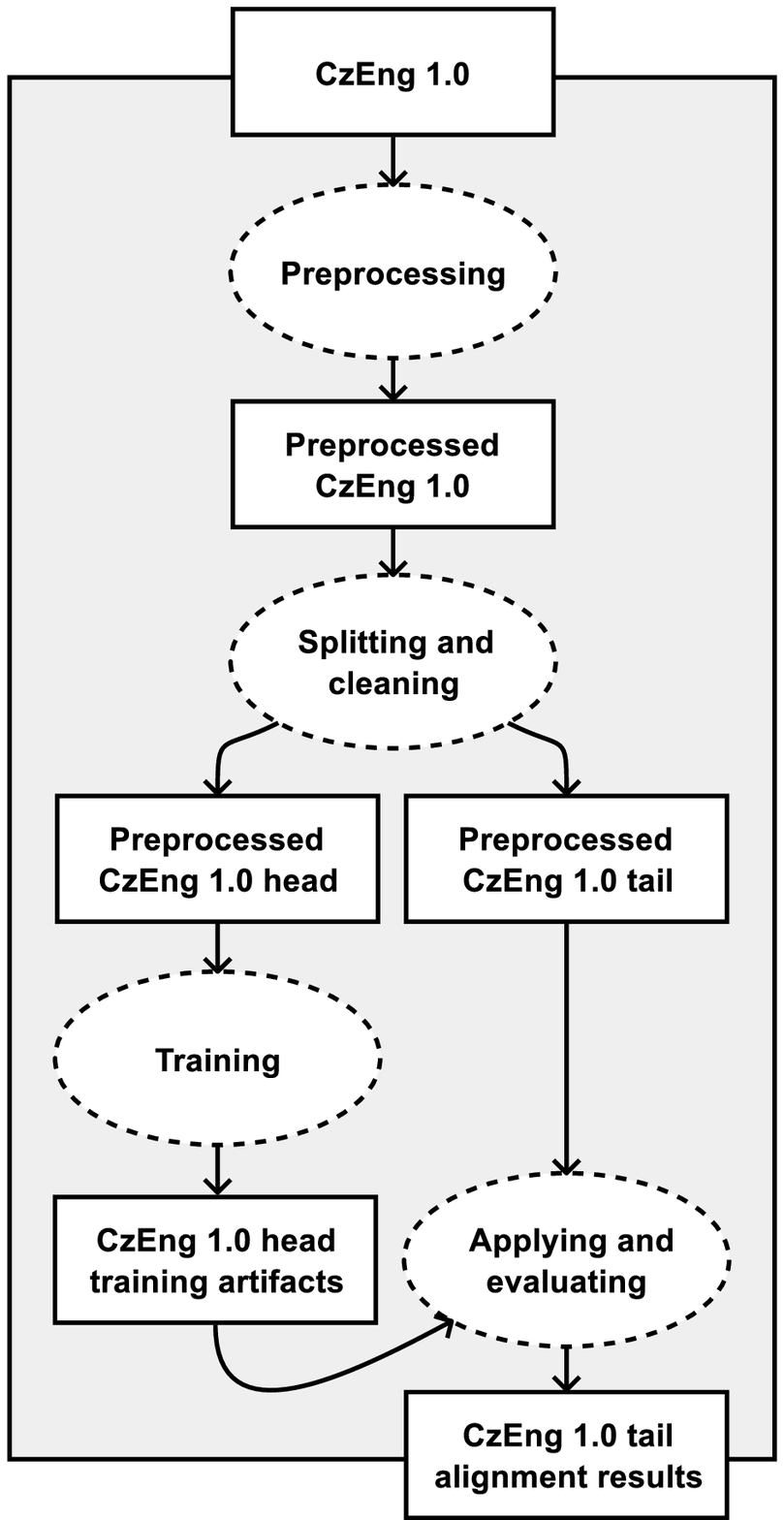}
    \captionsetup{justification=centering}
    \caption{Prealigned Data (CzEng 1.0) Experiment}
    \label{fig:experiment_1}
  \end{minipage}
  \space
  \begin{minipage}[b]{0.35\textwidth}
    \includegraphics[width=\textwidth]{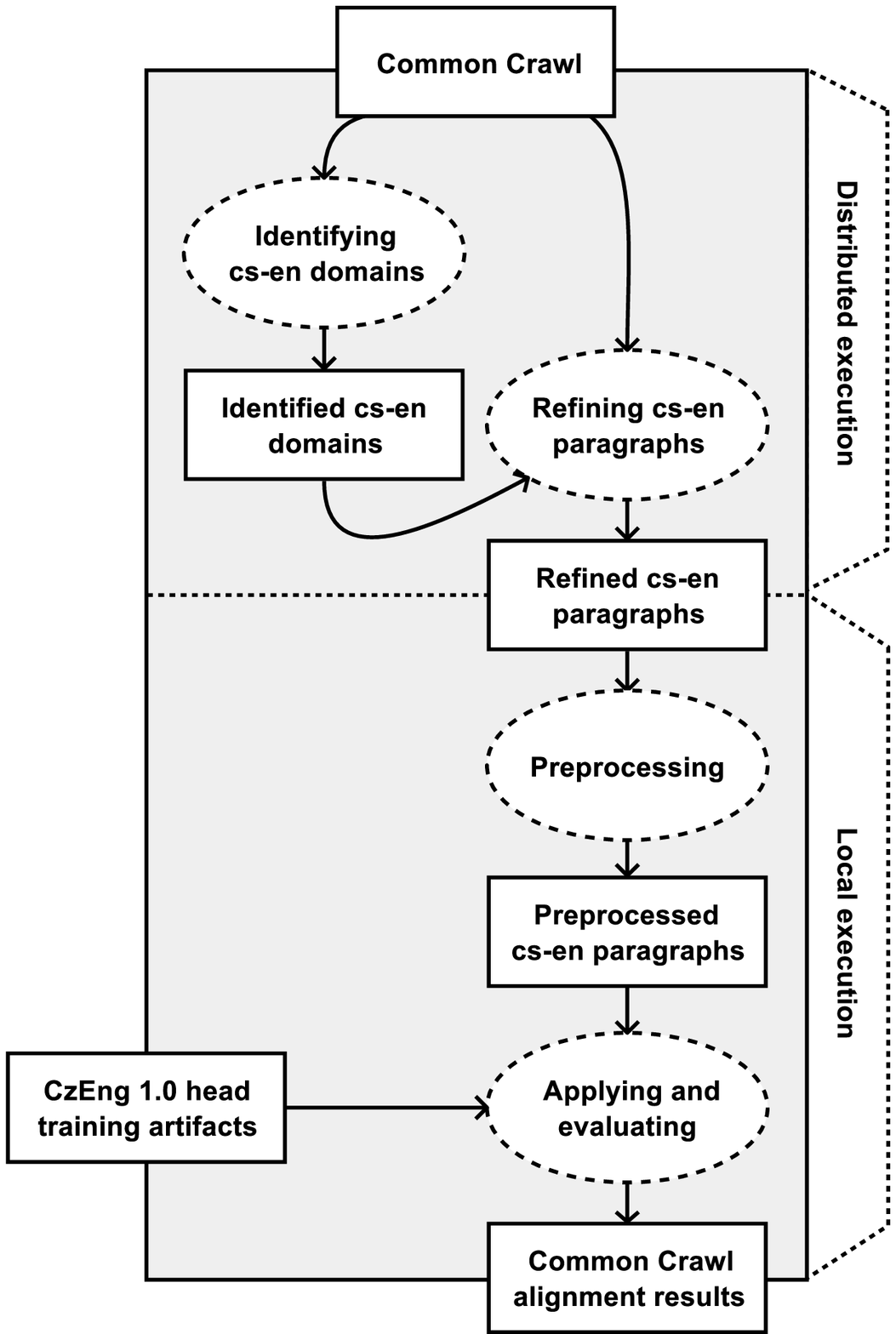}
    \captionsetup{justification=centering}
    \caption{Web Data (Common~Crawl) Experiment}
    \label{fig:experiment_2}
  \end{minipage}
\end{figure}

The corpus consists of all the training sections (packs 00--97) of CzEng 1.0 in the plain text, untokenized format. It
includes $14,833,358$ pairs of parallel sentences collected from various domains (e.g.~fiction, legislation, movie
subtitles, parallel web pages, etc.). By default, the pairs are shuffled, meaning they are not grouped by their domains.
The shuffled corpus is split exactly in half into a \textit{head} (i.e.~seed corpus) for training and a \textit{tail}
for evaluation. The whole procedure is illustrated in Figure~\ref{fig:experiment_1}.

The preprocessing step consists of tokenization and lowercasing. The head is cleaned by excluding all such pairs where
one of the sentences contains too many tokens or does not contain any letter from any alphabet. The tail is cleaned by
only applying the latter of the two mentioned criteria. The pairs containing overly long sentences are removed from the
head to get better quality word alignment.

During the training, SyMGIZA++ uses the union method for the final symmetrization of the word alignments, while bivec
uses the bilingual skip-gram model and runs for $10$ iterations. The pairs of parallel sentences from the head are
distributed into a set of artificial bins to form a dataset for the training of the classifier. Each bin contains
$50,000$ pairs of parallel documents from various domains, i.e.~$100,000$ individual documents. As discussed above, we
believe this amount of documents to be a good upper-bound estimate of the total number of paragraphs in either of the
two languages (Czech and English) located on a typical Czech--English web domain. Annoy builds search indexes with $500$
trees and during each search it inspects up to $20,000$ nodes to return $20$ candidates. The classifier is trained using
approximately $20\%$ of all the available pairs of Czech documents with their corresponding top English candidates.
Additionally, the supervised dataset contains nearly as many parallel pairs as non-parallel. The classifier's network is
trained for $20$ epochs with $1\%$ learning rate.

The trained method is used to realign the tail. The pairs of parallel sentences from the tail are distributed into
artificial bins in the same manner as those from the head. In this scenario, each bin simulates a web domain with
$50,000$ Czech and $50,000$ English paragraphs that we want to align. In contrast to real-world websites, these are
perfectly parallel, meaning that all the content is available in both languages. The original alignment of the tail is
forgotten and it does not affect the evaluation process is any way. The confidence threshold of the classifier is set to
$50\%$. The refined alignments represent a subset of all the pairs of Czech documents with their top English candidates
that the classifier accepts to be parallel.

In the preliminary alignments of the tail, $50.30\%$ of all the top candidates are exact matches. However, this ratio is
more satisfactory in the scored alignments---$71.30\%$. This means that the scoring of the preliminary alignments is an
important step in the whole process. The overall effectiveness of our method (after the application of the binary
classifier), when realigning the tail part of CzEng 1.0, is listed in Table~\ref{tab:czeng_effectiveness}. Of all the
existing pairs of parallel sentences $63.02\%$ were detected, and $93.74\%$ of all the detected pairs were correct.

\begin{table}[!htb]\centering
  \begin{tabular}{lr}
    \toprule
    \textbf{Recall (\%)} & 63.02 \\
    \textbf{Precision (\%)} & 93.74 \\
    \bottomrule
  \end{tabular}
  \caption{Prealigned Data (CzEng 1.0); Experiment: Effectiveness}
  \label{tab:czeng_effectiveness}
\end{table}

The computer used for the execution has Intel\textregistered{} Xeon\textregistered{} CPU E5-2630 v3 (20 MB Cache, 2.40
GHz) and 128 GB of memory. Table~\ref{tab:czeng_time} lists the approximate time durations of the individual steps of
the experiment.

\begin{table}[!htb]\centering
	\begin{tabular}{lr}
		\toprule
		\multicolumn{1}{c}{\textbf{Activity}} & \multicolumn{1}{c}{\textbf{Duration (hh:mm)}} \\
		\midrule
		\textbf{Preprocessing} \\
		Tokenizing and lowercasing & 00:08 \\
		Splitting and cleaning & 00:05 \\
		\midrule
		\textbf{Training part I} \\
		Applying SyMGIZA++ & 13:21 \\
		Generating dictionary & 00:10 \\
		Applying bivec & 01:01 \\
		\midrule
		\textbf{Training part II} \\
		Generating document vectors & 00:37 \\
		Aligning document vectors (Annoy) & 05:52 \\
		Scoring alignments & 02:49 \\
		Training binary classifier & 01:29 \\
		\midrule
		\textbf{Application} \\
		Generating document vectors & 00:45 \\
		Aligning document vectors (Annoy) & 07:04 \\
		Scoring alignments & 04:10 \\
		Applying binary classifier & 00:47 \\
		\bottomrule
	\end{tabular}
	\caption{Prealigned Data (CzEng 1.0); Experiment: Time Duration}
	\label{tab:czeng_time}
\end{table}

\subsection{Web Data (Common Crawl) Experiment}
\label{sec:web_data_experiment}

The second experiment deals with the non-parallel, real-word, noisy data acquired from the web. The language pair of our
interest is again Czech--English. The procedure uses the training artifacts created in the first experiment, namely the
dictionary, bilingual word vectors and the trained classifier.

The input data are obtained from the July 2015 dataset provided by Common Crawl Foundation and consist of approximately
$1.84$ billions of crawled web pages, taking about $149$ TB of disk space in an uncompressed format. To store and
process this large volume of data we use Hadoop---an HDFS~\cite{Shvachko:2010} cluster and the
MapReduce~\cite{Dean:2004} framework.

The procedure of the experiment is illustrated in Figure~\ref{fig:experiment_2}. It starts with a distributed execution
running two MapReduce jobs. The first job creates a list of web domains containing at least some Czech and English
paragraphs, i.e.~contents of~\texttt{<p>} HTML tags. Parsing of the HTML is done using
jsoup\footnote{\url{https://jsoup.org/}} and language detection is performed by
language-detector.\footnote{\url{https://github.com/optimaize/language-detector}} Due to the unsatisfactory
effectiveness of language-detector on shorter texts, paragraphs having less than $100$ characters are discarded. In the
future, the parsing could be modified to consider also other HTML tags.

The list of web domains is further filtered to keep only those having the ratio of Czech to English paragraphs within
the interval $(0.01, 100)$. This filtering discards all domains with very unbalanced language distribution. The
output of the first job contains $8,750$ identified bilingual domains.

The second MapReduce job extracts the Czech and English paragraphs for all the identified bilingual web domains. In
order to provide the second job with the file containing the accepted domains, Hadoop Distributed Cache is utilized. The
output of the second job contains $5,931,091$ paragraphs for both languages, namely $801,116$ Czech and $5,129,975$
English. These paragraphs are aligned with our method in a local execution and the results are evaluated. All the
settings remain the same as in the first experiment, except the threshold of the classifier is changed to $99\%$. The
precision is favored over the recall. The extracted paragraph-aligned parallel corpus contains $114,771$ pairs from
$2,178$ domains, having in total $7,235,908$ Czech and $8,369,870$ English tokens. Table~\ref{tab:common_crawl_domains}
lists the most frequent web domains contributing to the extracted corpus. The size of the extracted corpus is comparable
with the amount of Czech--English parallel data acquired by the related project focused on mining the Common Crawl
datasets~\cite{Smith:2013}.

The quality of the extracted corpus is evaluated manually on a set of $500$ randomly selected paragraph pairs. The
inspected pairs are categorized into the categories displayed in Table~\ref{tab:common_crawl_500}. A pair of paragraphs
is considered to be a human translation if it seems like created by a human. If the translation of the pair seems
cumbersome, it is labeled as a product of machine translation. A partial match represents a situation, when one
paragraph is incomplete regarding the content of the other one. Everything else is labeled as a mismatch. If we consider
the pairs belonging to the two categories of human and machine translation as true positives, then the estimation of
precision is $94.60\%$.

\begin{table}[!htb]\centering
	\begin{tabular}{lrr}
		\toprule
		\textbf{Source Domain} & \textbf{Paragraph Pairs} & \textbf{Ratio (\%)} \\
		\midrule
		europa.eu & 23457 & 20.45 \\
		eur-lex.europa.eu & 15037 & 13.11 \\
		windows.microsoft.com & 11905 & 10.38 \\
		www.europarl.europa.eu & 8560 & 7.46 \\
		www.project-syndicate.org & 2210 & 1.93 \\
		www.debian.org & 2191 & 1.91 \\
		support.office.com & 1908 & 1.66 \\
		www.esa.int & 1308 & 1.14 \\
		www.eea.europa.eu & 1299 & 1.13 \\
		www.muni.cz & 1206 & 1.05 \\
		\vdots & \vdots & \vdots  \\
		\midrule
		\textbf{Total} & 114,711 & 100.00 \\
		\bottomrule
	\end{tabular}
	\caption{Web Data (Common Crawl); Experiment: Web Domains}
	\label{tab:common_crawl_domains}
\end{table}

\begin{table}[!htb]\centering
  \begin{tabular}{lrr}
    \toprule
    \textbf{Category} & \textbf{Count} & \textbf{Ratio (\%)} \\
    \midrule
    Human translation & 466 & 93.20 \\
    Machine translation & 7 & 1.40 \\
    Partial match & 13 & 2.60 \\
    Mismatch & 14 & 2.80 \\
    \midrule
    \textbf{Total} & 500 & 100.00 \\
    \bottomrule
  \end{tabular}
  \caption{Web Data (Common Crawl); Experiment: Evaluation (500 Paragraph Pairs)}
  \label{tab:common_crawl_500}
\end{table}

Manual evaluation of the method's recall is complicated and therefore it is performed using only one selected web
domain---\texttt{www.csa.cz}, the official website of Czech Airlines. The input dataset contains $68$ Czech and $87$
English paragraphs for this domain. These paragraphs are manually aligned, creating the \textit{desirable alignments}.
When evaluated, the desirable alignments contain $44$ paragraph pairs, of which $42$ also appear in the corpus extracted
by our method. Additionally, the extracted corpus include $1$ extra pair subjectively regarded as a mismatch.
Table~\ref{tab:common_crawl_csa} shows the effectiveness of our method evaluated for the \texttt{www.csa.cz} web domain.

\begin{table}[!htb]\centering
  \begin{tabular}{lr}
    \toprule
    \textbf{Recall (\%)} & 95.45 \\
    \textbf{Precision (\%)} & 97.67 \\
    \bottomrule
  \end{tabular}
  \caption{Web Data (Common Crawl)\\Experiment: Effectiveness (\texttt{www.csa.cz})}
  \label{tab:common_crawl_csa}
\end{table}

The Hadoop cluster used for the distributed execution of the two MapReduce jobs consists of 3 management nodes and 24
worker nodes. The management nodes run components like front-end, HDFS NameNode and MapReduce History Server. Every node
of the configuration has Intel\textregistered{} Xeon\textregistered{} CPU E5-2630 v3 (20 MB Cache, 2.40 GHz) and 128 GB
of memory. The total disk space available on the cluster is 1.02 PB. The HDFS operates with a replication factor of 4.
The rest of the procedure, i.e.~the local execution, is done on one node of the cluster.
Table~\ref{tab:common_crawl_time} contains the approximate time durations of the individual steps of the experiment.

\begin{table}[!htb]\centering
	\begin{tabular}{lr}
		\toprule
		\multicolumn{1}{c}{\textbf{Activity}} & \multicolumn{1}{c}{\textbf{Duration (hh:mm)}} \\
		\midrule
		\textbf{MapReduce framework} \\
		Identifying cs-en domains & 11:58 \\
		Refining cs-en paragraphs & 11:38 \\
		\midrule
		\textbf{Local Execution} \\
		Tokenization and lowercasing & 00:09 \\
		Generating document vectors & 00:58 \\
		Aligning document vectors (Annoy) & 01:13 \\
		Scoring alignments & 03:42 \\
		Applying classifier & 00:39 \\
		\bottomrule
	\end{tabular}
	\caption{Web Data (Common Crawl); Experiment: Time Duration}
	\label{tab:common_crawl_time}
\end{table}

\section{Conclusions and Future Work}
\label{sec:conclusion_and_future_work}

The majority of methods for bilingual document alignment search for pairs of parallel web pages by comparing the
similarity of their HTML structures. Our method does not depend on any kind of page structure comparison. We are able to
efficiently identify pairs of parallel segments (i.e.~paragraphs) located anywhere on the pages of a web domain,
regardless of their structure.

To verify the idea of our method, we have performed two experiments focused on the Czech--English language pair with
both prealigned and real-world data. These experiments show satisfactory results, implying that the proposed method is a
promising baseline for acquiring parallel corpora from the web.

Nevertheless, there is still some room for improvement. First of all, our method does not consider word order at any
stage during the aligning process. The scoring function and the features of the classifier could be extended to take
word order into account. Then, there is an asymmetric nature of our method, meaning that it yields different results if
the source and the target languages are swapped. The method could perform the alignment for both directions and the
results could be symmetrized. This might help to achieve an even higher precision.

Another possibility would be to extend our method with some kind of structural comparison, for instance, in form of a
new feature for the classifier, that would compare the structural origin of the input documents (e.g.~XPath
of~\texttt{<p>} tags, in case of aligning paragraphs from the web).

Finally, we have used our method only in a single-node environment so far. This is largely because we have worked with
relatively small sets of documents (not more than $15,000,000$). However, the method is designed to be able to run in
distributed fashion. Bins with input documents represent independent and isolable tasks. With the method trained in a
local execution, these tasks could be distributed across multiple nodes of a cluster. This could increase the throughput
of our method, and hence decrease the overall execution time.

\section*{Acknowledgements}

Access to computing and storage facilities owned by parties and projects contributing to the National Grid
Infrastructure MetaCentrum, provided under the programme "Projects of Large Research, Development, and Innovations
Infrastructures" (CESNET LM2015042), is greatly appreciated.

This work has received funding from the European Union's Horizon 2020 research and innovation programme under grant
agreement no. 644402 (HimL). This work has been using language resources stored and distributed by the LINDAT/CLARIN
project of the Ministry of Education, Youth and Sports of the Czech Republic (project LM2015071).

\bibliography{mybib}

\correspondingaddress
\end{document}